\documentclass[letterpaper]{article} 
\usepackage{aaai2026}
\usepackage{times}  
\usepackage{helvet}  
\usepackage{courier}  
\usepackage[hyphens]{url}  
\usepackage{graphicx} 
\usepackage{natbib}  
\usepackage{caption} 
\usepackage{algorithm}
\usepackage{algorithmic}

\usepackage{paralist}

\usepackage{newfloat}
\usepackage{listings}
\DeclareCaptionStyle{ruled}{labelfont=normalfont,labelsep=colon,strut=off} 
\floatstyle{ruled}
\newfloat{listing}{tb}{lst}{}
\floatname{listing}{Listing}
\usepackage{xcolor}
\usepackage{paralist}

\title{Fixed-Persona SLMs with Modular Memory: Scalable NPC Dialogue on Consumer Hardware}
\author{
    Martin Braas Andreasen\textsuperscript{\rm 1},
    Lukas Esterle\textsuperscript{\rm 1,\rm 2}
}
\affiliations{
    \textsuperscript{\rm 1} Department of Electrical and Computer Engineering, Aarhus University\\
    \textsuperscript{\rm 2} DIGIT\\

    mba@ece.au.dk; lukas.esterle@ece.au.dk
}

\usepackage{bibentry}

\begin{document}

\maketitle

\begin{abstract}
Large Language Models (LLMs) have demonstrated remarkable capabilities in generating human-like text, yet their applicability to dialogue systems in computer games remains limited. This limitation arises from their substantial hardware requirements, latency constraints, and the necessity to maintain clearly defined knowledge boundaries within a game setting. In this paper, we propose a modular NPC dialogue system that leverages Small Language Models (SLMs), fine-tuned to encode specific NPC personas and integrated with runtime-swappable memory modules. These memory modules preserve character-specific conversational context and world knowledge, enabling expressive interactions and long-term memory without retraining or model reloading during gameplay. We comprehensively evaluate our system using three open-source SLMs: DistilGPT-2, TinyLlama-1.1B-Chat, and Mistral-7B-Instruct, trained on synthetic persona-aligned data and benchmarked on consumer-grade hardware. While our approach is motivated by applications in gaming, its modular design and persona-driven memory architecture hold significant potential for broader adoption in domains requiring expressive, scalable, and memory-rich conversational agents, such as virtual assistants, customer support bots, or interactive educational systems.
\end{abstract}


\section{Introduction}
\label{sec:intro}

Non-playable characters (NPCs) are central to immersive storytelling and dynamic player experiences in modern video games. Yet, most commercial NPC systems rely on handcrafted dialogue trees or rigid behavior scripts, which scale poorly and fail to produce nuanced or long-term coherent interactions \cite{schlunder2013greetings}. More recently, large language models (LLMs) have enabled open-ended, expressive dialogue generation, but their substantial hardware requirements and reliance on cloud APIs make them impractical for local, real-time use in games, especially when multiple distinct characters are needed \cite{sarahparvinicanaimakegamesmoreimmersive, gao-emami-2023-turing}. 

In this work, we explore an alternative design: Small Language Models (SLMs) fine-tuned for specific NPC personas, paired with runtime-swappable memory modules that preserve character-specific knowledge and conversational context. At runtime, each model uses two distinct types of memory: conversational memory, preserving previous interactions between the NPC and player to maintain continuity, and world knowledge memory, providing contextually relevant facts and narrative background specific to the NPC's role. This architecture enables the deployment of multiple, expressive NPCs with long-term memory without retraining or reloading new models during gameplay. Each SLM is trained with a distinct, embedded persona using a lightweight fine-tuning pipeline. At runtime, the model is coupled with a dedicated memory store, dynamically swappable to reflect individual NPC instances in the game

This design preserves character identity (persona is fixed in training), while supporting flexible, context-rich interactions through memory. It allows the same model to power multiple NPCs of the same type, such as several innkeepers or guards, pilots or mechanics, each with their own unique memory and interaction history. The result is a scalable, fully local NPC dialogue system, meaning that models can run directly on consumer hardware without relying on external cloud services or APIs, delivering high quality responses even under hardware constraints. Although presented primarily in the context of NPCs and computer games, its modular design and persona-driven approach make it broadly applicable to any domain requiring expressive, memory-rich conversational agents, such as virtual assistants, customer support bots, or interactive educational systems.

To support this architecture, we implement a full pipeline: a multi-stage fine-tuning process for fast NPC creation, a modular memory system using ChromaDB, and (currently, pre-game integration) a CLI-based runtime that composes prompts dynamically with retrieved memory and player input. We evaluate our system using three open-source SLMs: DistilGPT-2~\cite{distilgpt2}, TinyLlama-1.1B-Chat~\cite{zhang2024tinyllamaopensourcesmalllanguage, tinyllama2023}, and Mistral-7B-Instruct~\cite{jiang2023mistral7b, mistral2023}, all trained on synthetic persona-aligned data, and benchmarked on consumer-grade hardware.

This paper makes the following contributions:

\begin{compactitem}
    \item \textbf{A modular NPC dialogue system} where each fine-tuned SLM encodes a fixed persona, and memory modules can be swapped at runtime to enable distinct character instances.
    \item \textbf{A lightweight seed-based fine-tuning pipeline} that enables rapid persona instantiation without relying on proprietary models.
    \item \textbf{A retrieval-augmented runtime framework} for managing conversational history and world knowledge in memory, supporting long-term coherent interactions.
    \item \textbf{A comprehensive evaluation} across factuality, memory retention, fluency, latency, and memory usage - demonstrating the feasibility of expressive NPC dialogue on local consumer hardware.
\end{compactitem}
\section{Related Work}
\label{sec:relWork}

Dialogue systems for non-playable characters (NPCs) have historically relied on handcrafted dialogue trees or rule-based methods, which limit scalability and expressive interaction~\cite{schlunder2013greetings}. Recent interest has shifted towards leveraging generative AI models, such as transformer-based architectures, to enhance NPC realism and interactivity in games~\cite{gao-emami-2023-turing, sarahparvinicanaimakegamesmoreimmersive}. NVIDIA's ACE platform exemplifies recent industry interest in using generative AI for expressive, interactive NPC dialogues~\cite{nvidia2023ace}.

Efficient deployment of dialogue models remains challenging due to computational constraints. Approaches such as distillation, exemplified by DistilBERT and DistilGPT-2, have shown how models can retain substantial performance while significantly reducing computational demands~\cite{sanh2020distilbertdistilledversionbert, distilgpt2}. Similarly, the development of small but performant open-source language models, such as TinyLlama~\cite{zhang2024tinyllamaopensourcesmalllanguage,tinyllama2023}, and Mistral~\cite{jiang2023mistral7b}, illustrates ongoing efforts to balance efficiency and expressivity. Recent industry applications also confirm the growing viability of small language models (SLMs) for on-device game character dialogues~\cite{ikennoli2024deploythefirstondeviceslmforimprovedgamecharacterroleplay}.

Parameter-efficient fine-tuning methods~\cite{houlsby2019parameterefficienttransferlearningnlp} have enabled further reductions in computational overhead. Techniques such as LoRA (Low-Rank Adaptation~\cite{hu2021lora}) significantly decrease the number of trainable parameters, making fine-tuning on limited resources feasible without sacrificing substantial performance.

Our approach uniquely integrates and extends these strands of research. Specifically, we demonstrate how combining small language models, efficient LoRA fine-tuning, and runtime-swappable modular memory provides a scalable and resource-efficient architecture specifically tailored for interactive NPC dialogue systems, but potentially applicable to other conversational scenarios as well.
\section{System Overview: Fixed-Persona, Modular-Memory NPCs}
\label{sec:sysOverview}

 The proposed system enables scalable deployment of non-playable characters (NPCs) using small language models (SLMs) that are fine-tuned with fixed personas and coupled with modular memory components at runtime. This separation between character identity and dialogue context allows a single model to support multiple, memory-rich NPC instances without retraining or modification.
%
 At a high level, each NPC model serves as a persona-anchored dialogue agent. The model is responsible for maintaining in-character behavior, tone and knowledge boundaries. During deployment, the system combines this static behavioral core with dynamic memory components to generate contextually appropriate and temporally coherent dialogue. 

 The full system includes:
\begin{inparaenum}[(i)]
    \item the fine-tuned SLM backend;
    \item two modular memory stores per NPC;
    \item a runtime prompt composer, and
    \item a command-line interface for interaction and evaluation.
\end{inparaenum}

\subsection{Fixed Persona via Fine-Tuning}
Each NPC's personality is fine-tuned using Low-Rank Adaptation (LoRA), training on a small curated dataset. 
This dataset encodes the character's voice, behavior constraints, and permissible knowledge. For example, a friendly innkeeper NPC is trained to avoid topics beyond their world knowledge or behave outside their temperament.

These personas are not injected 
at runtime. The result is a stable, character-consistent response style that avoids the brittleness and token overhead of in-context prompt conditioning. While the persona remains static, dynamic variability is introduced via the modular memory system.

\subsection{Modular Memory Store (Runtime-Swappable)}

To enable unique, context-aware interactions for each NPC instance, the system employs two separate vector memory stores using ChromaDB~\cite{chromadb2023}:

\begin{itemize}
    \item \textbf{Conversation memory} stores prior interactions between the player and the NPC. This supports long-term continuity, personal familiarity, and contextual grounding.
    \item \textbf{World knowledge} stores structured facts, background information, or narrative hooks relevant to the NPC's domain or role.
\end{itemize}

At runtime, these memory stores are queried independently using cosine similarity between the incoming player input and stored embeddings~\cite{reimers2019sentencebertsentenceembeddingsusing}. The system retrieves the top-$k$ relevant entries from each store and integrates them into the prompt.
Because the memory is modular and decoupled from the model, different NPC instances can share the same SLM while maintaining distinct histories and knowledge bases. This allows, for example, multiple innkeeper characters across the game world to be powered by a single innkeeper model while exhibiting different contextual knowledge and memory.

\subsection{Runtime Dialogue Pipeline}

The system's dialogue generation process follows a structured pipeline as seen in Figure \ref{runtime-dialogue-pipeline-figure}:

\begin{figure}[h]
\centering
\includegraphics[width=.7\columnwidth]{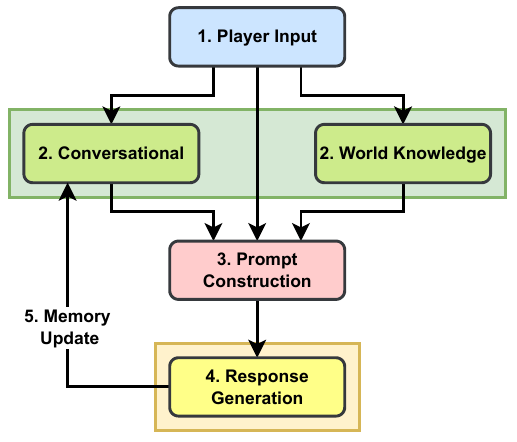}
\caption{Diagram depicting the Runtime Dialogue Pipeline}
\label{runtime-dialogue-pipeline-figure}
\end{figure}

\begin{enumerate}
    \item \textbf{Player Input:} A prompt is entered by the player via the command-line interface (or in the game, after integration)
    \item \textbf{Memory Retrieval:} The system searches both the conversational and world knowledge memory stores for the top-$k$ most relevant entries.
    \item \textbf{Prompt Construction:} Retrieved entries are concatenated with the player's input and passed through a formatting layer, producing the final model prompt.
    \item \textbf{Response Generation:} The SLM, guided by its fixed persona, and augmented by contextual memory, generates a character-consistent response.
    \item \textbf{Memory Update:} The new dialogue turn is appended to the conversation memory for future reference.
\end{enumerate}

This pipeline supports fast interaction and dynamic memory use without requiring the model to be reloaded, modified, or recompiled. The results is a scalable and expressive NPC dialogue system capable of running entirely on consumer-grade hardware.
\section{Fine-tuning pipeline}
\label{sec:fine-tuning}

To support character-consistent dialogue generation, each NPC model in the system is fine-tuned to encode a fixed persona aligned with its narrative role. This section outlines the lightweight, multi-stage fine-tuning pipeline used to create expressive NPCs from open-source small language models (SLMs), while maintaining compatibility with consumer-grade hardware.

The process begins with the creation of a small seed dataset consisting of 10-20 handcrafted prompt-response pairs. These examples are written to reflect the intended persona's voice, tone, and domain-specific behavior. For instance, an innkeeper character may answer questions about local inns and travelers, while avoiding topics unrelated to their knowledge scope.

This seed dataset is used to perform a preliminary LoRA fine-tuning of an intermediate model, which serves as a synthetic data generator~\cite{hu2021lora}. The intermediate model is prompted with new player inputs to generate a larger set of synthetic prompt-response pairs, typically around 150 entries but can be extended depending on the desired training depth. These generations are manually reviewed to ensure alignment with the intended persona and consistency with in-universe knowledge constraints.

This dataset is then used to fine-tune the NPCs, but the process can be repeated to generate even larger synthetic dataset. When using LoRA, great results have been achieved using "smaller" datasets, but the approach can be repeated to create larger synthetic datasets, which may be used to perform full fine-tuning. This "staged" approach enables high-quality persona alignment while minimizing manual annotation effort, and avoids reliance on large proprietary models during data generation.

This pipeline was applied to generate a dataset for a merchant-persona which was used on all three base-models to enable easy comparison of performance. First, we generated an initial curated dataset ($\sim$115 pairs, after manual review) to fine-tune an intermediate NPC model, which subsequently produced a larger synthetic data ($\sim$564 pairs). This was done to explore the impact of different dataset sizes when using LoRA and the different base-models. This led to the creation of 7 models, named after base-model and training dataset size. The naming convention can be seen in table \ref{tab:npc-naming-convention}

\begin{table}[h]
    \centering
    \begin{tabular}{|c|c|c|}
        \hline
         \textbf{Name} & \textbf{Base Model} & \textbf{Dataset Size} \\
         \hline
         JackS & DistilGPT-2 & Small \\
         JackL & DistilGPT-2 & Large \\
         \hline
         CasperS & TinyLlama-1.1B-Chat & Small \\
         CasperL & TinyLlama-1.1B-Chat & Large \\
         \hline
         OliverS & Mistral-7B-Instruct & Small \\
         OliverL & Mistral-7B-Instruct & Large \\
         OliverQ & Mistral-7B-Instruct & Small \\
         \hline
    \end{tabular}
    \caption{NPC's naming conventions}
    \label{tab:npc-naming-convention}
\end{table}

The NPCs are assigned development names, with a suffix annotating whether they have been trained on the large ("L", e.g., OliverL) or small ("S", e.g., OliverS) dataset. The only exemption from this naming convention is OliverQ, which was created to explore the impact of quantization using AutoGTPQ~\cite{frantar2023gptqaccurateposttrainingquantization}.

The result of this pipeline is a set of fixed-persona NPCs that maintain consistent behavioral boundaries and tone, while being lightweight enough to deploy locally alongside modular memory systems (as described in Section \ref{sec:sysOverview}). The next section details how these models are evaluated across quality and system performance dimensions.
\section{Results}
\label{sec:results}

This section presents the detailed results and discussions grouped into three main categories: Dialogue Quality, Hardware Efficiency, and Runtime Modularity. Each subsection includes a description of the evaluation methodology, visual representation of results, and interpretation. All experiments, including training and evaluation, has been conducted locally on a Windows 11 Desktop PC with an Intel Core i7-8700K CPU, 4x8GB 3200MHz RAM, and an NVIDIA RTX 2070 Super GPU (8GB VRAM). The system used Python 3.12.3 with Torch 2.6.0, Transformers 4.49.0, ChromaDB 0.6.3, and SentenceTransformers 3.4.1.

\subsection{Dialogue Quality}
\subsubsection{Factual Consistency}
We assessed NPC adherence to predefined persona knowledge using the Openchat-3.6 model as an automated judge~\cite{wang2024openchatadvancingopensourcelanguage}, prompt-engineered via an instruction template to evaluate factual correctness and appropriate refusal behavior (questions outside defined knowledge scope) across 100 responses per NPC variant.
\begin{figure}[h]
\centering
\includegraphics[width=.85\columnwidth, trim= 0 10 0 0,clip]{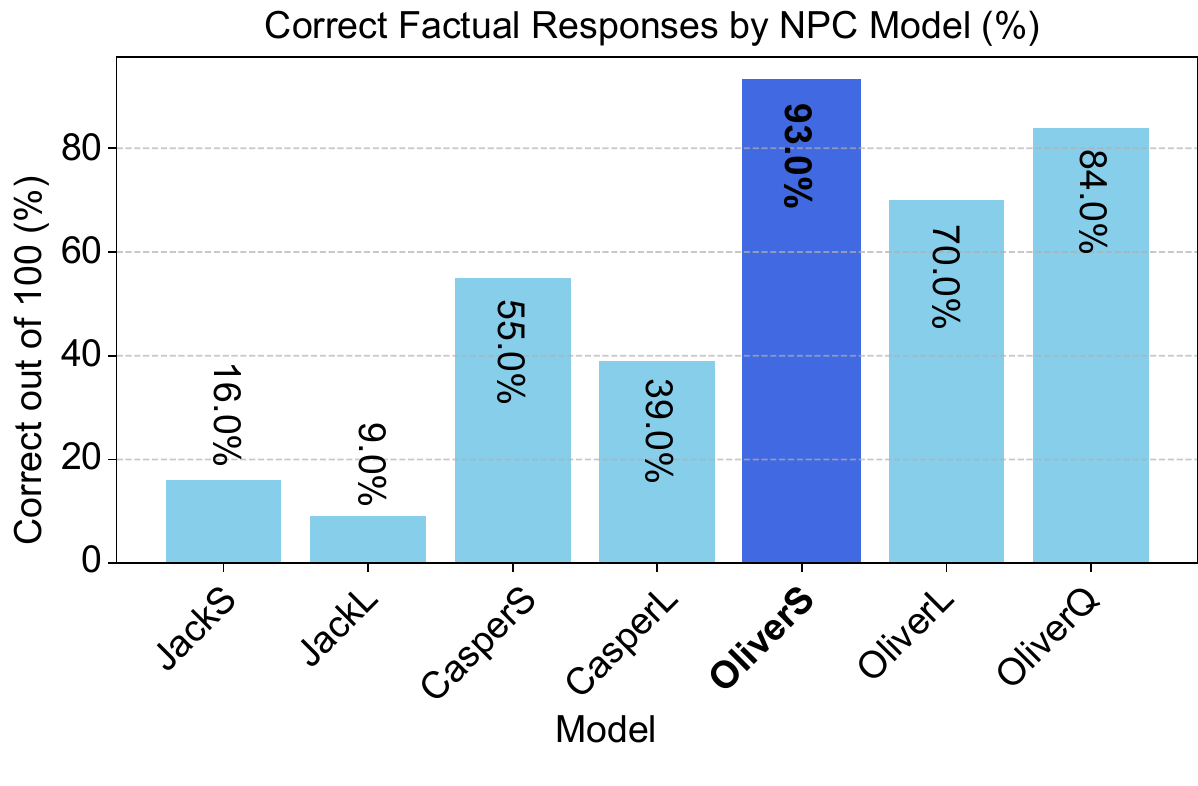}
\caption{Factual accuracy of NPC responses across different model variants.}
\label{factuality-test-diagram}
\end{figure}
Figure~\ref{factuality-test-diagram} show that OliverS significantly outperformed all other variants (93\%), and models with smaller datasets consistently performing better compared to the same models with larger datasets (JackS 16\% vs JackL 9\% and CasperS 55\% vs CasperL 39 \%), suggesting possible dataset quality issues or overfitting.

\subsubsection{Context Retention (Conversational Memory)}
We evaluated the NPC's ability to recall previously introduced information across multiple dialogue turns. In 30 multi-turn interactions, NPCs were tested on their ability to reference keywords introduced earlier, such as the player's name.

\begin{figure}[h]
\centering
\includegraphics[width=.8\columnwidth]{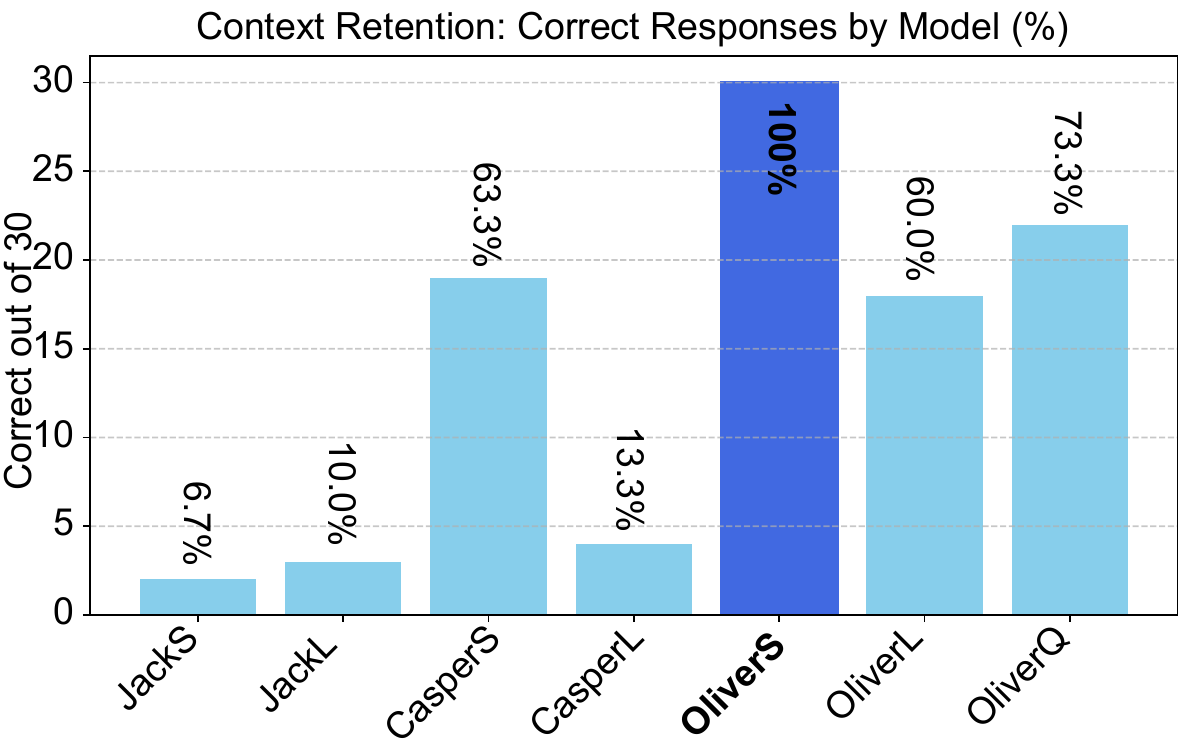} 
\caption{Context retention performance of each NPC variant, measured by the percentage of correct keyword recalls in multi-turn conversations.}
\label{context-retention-diagram}
\end{figure}

OliverS (100 \%) demonstrated perfect retention (Fig.~\ref{context-retention-diagram}), clearly outperforming other models, with CasperS (63.3 \%) showing moderate but surprisingly good results. Jack variants are notably lagging behind (JackS 6.7 \%, JackL 10 \%).

\subsubsection{World Knowledge Retrieval}
We measured NPC accuracy in retrieving specific information from their world knowledge database by systematically executing 30 queries per variant, each designed to trigger retrieval of distinct knowledge entries.
\begin{figure}[h]
\centering
\includegraphics[width=.8\columnwidth]{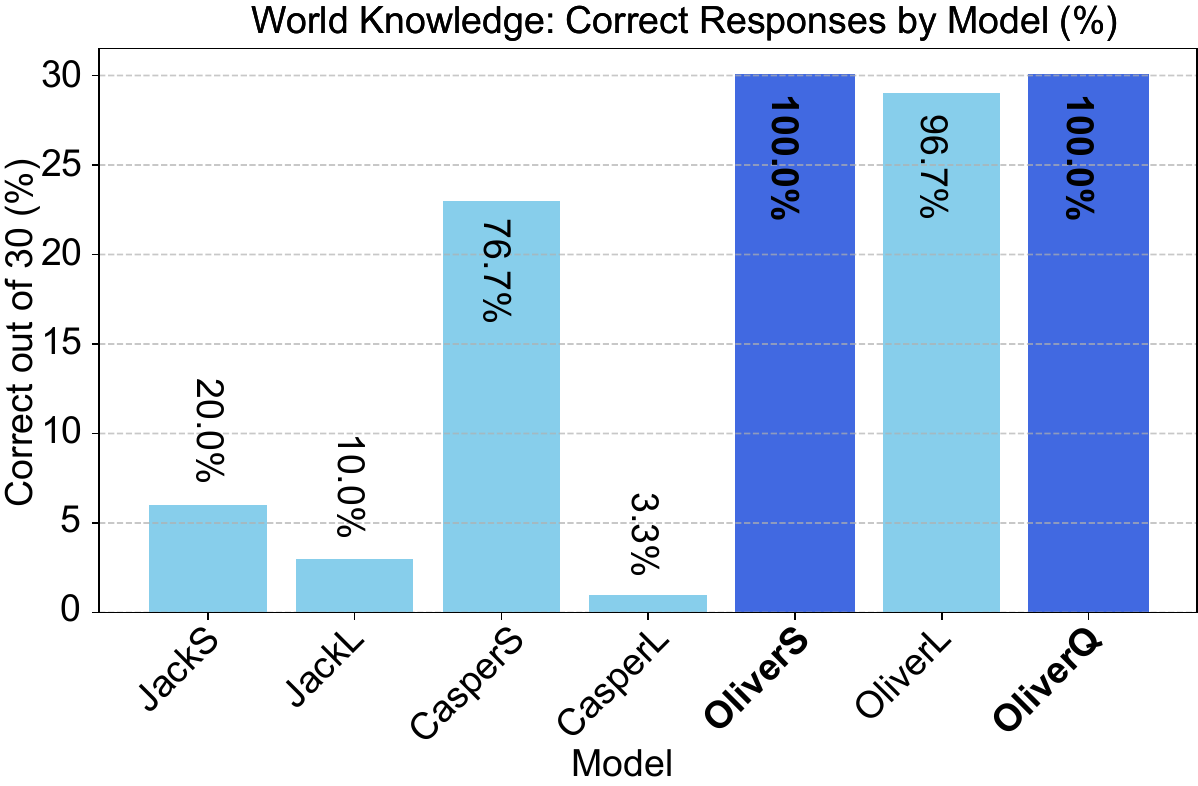} 
\caption{Accuracy of world knowledge retrieval by NPC model, illustrating the percentage of correctly retrieved entries from memory databases.}
\label{world-knowledge-diagram}
\end{figure}
Oliver variants (OliverS 100 \%, OliverL 96.7 \%, OliverQ 100 \%)  excelled with perfect and near-perfect retrieval (Fig.~\ref{world-knowledge-diagram}), while CasperS (76.7 \%) performed robustly, but CasperL and Jack variants showed significant shortcomings.

\subsubsection{Fluency (Grammar and Style)}
We evaluated linguistic fluency by analyzing grammatical, spelling, and stylistic errors in 30 NPC-generated responses per variant using the LanguageTool grammar checker~\cite{languagetool}. Results are reported as average errors per response.
\begin{figure}[h]
\centering
\includegraphics[width=.8\columnwidth]{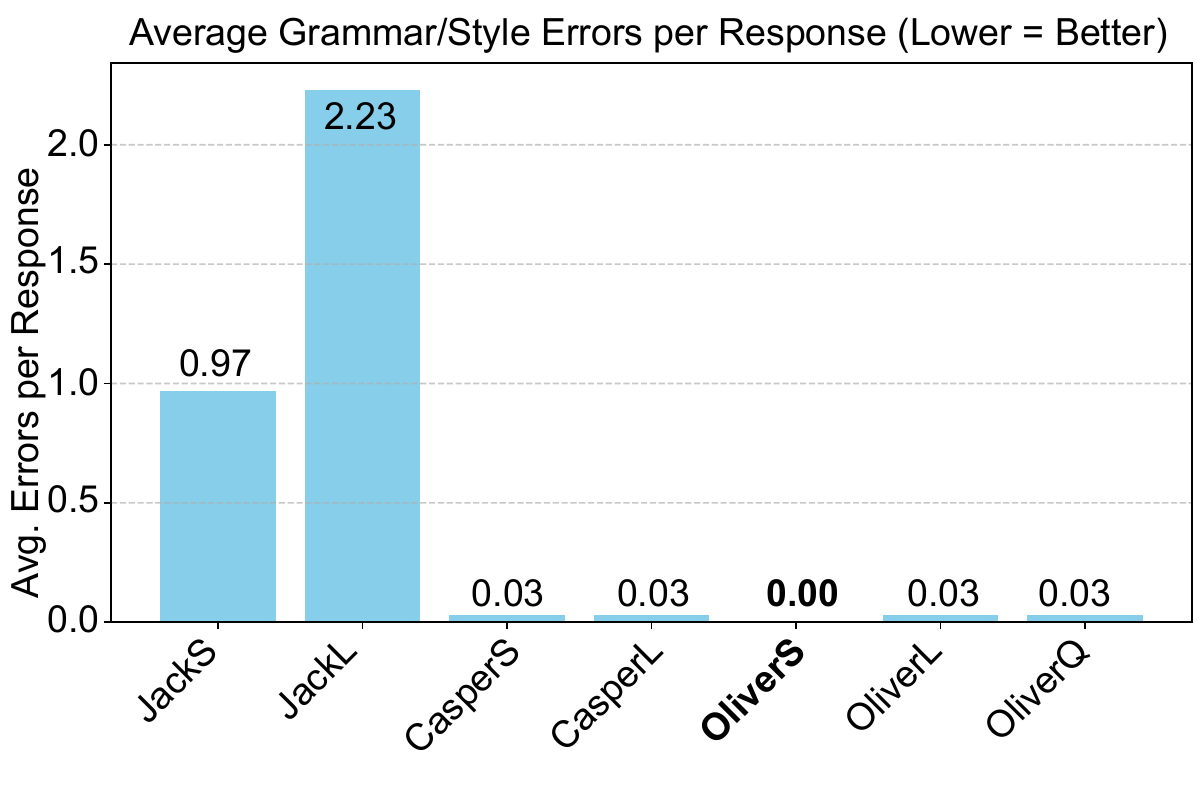} 
\caption{Average grammatical and stylistic errors per response produced by NPC models, as measured by LanguageTool. Lower values indicate better fluency.}
\label{language-tool-diagram}
\end{figure}
OliverS generated flawless responses (Fig.~\ref{language-tool-diagram}). Jack variants showed significant grammatical issues reflecting lower model complexity, while other models remained nearly error-free.

\subsection{Hardware Efficiency Metrics}
\subsubsection{GPU Memory Usage (VRAM)}
We measured mean GPU VRAM usage during inference, assessing suitability for local deployment.
\begin{figure}[h]
\centering
\includegraphics[width=.8\columnwidth]{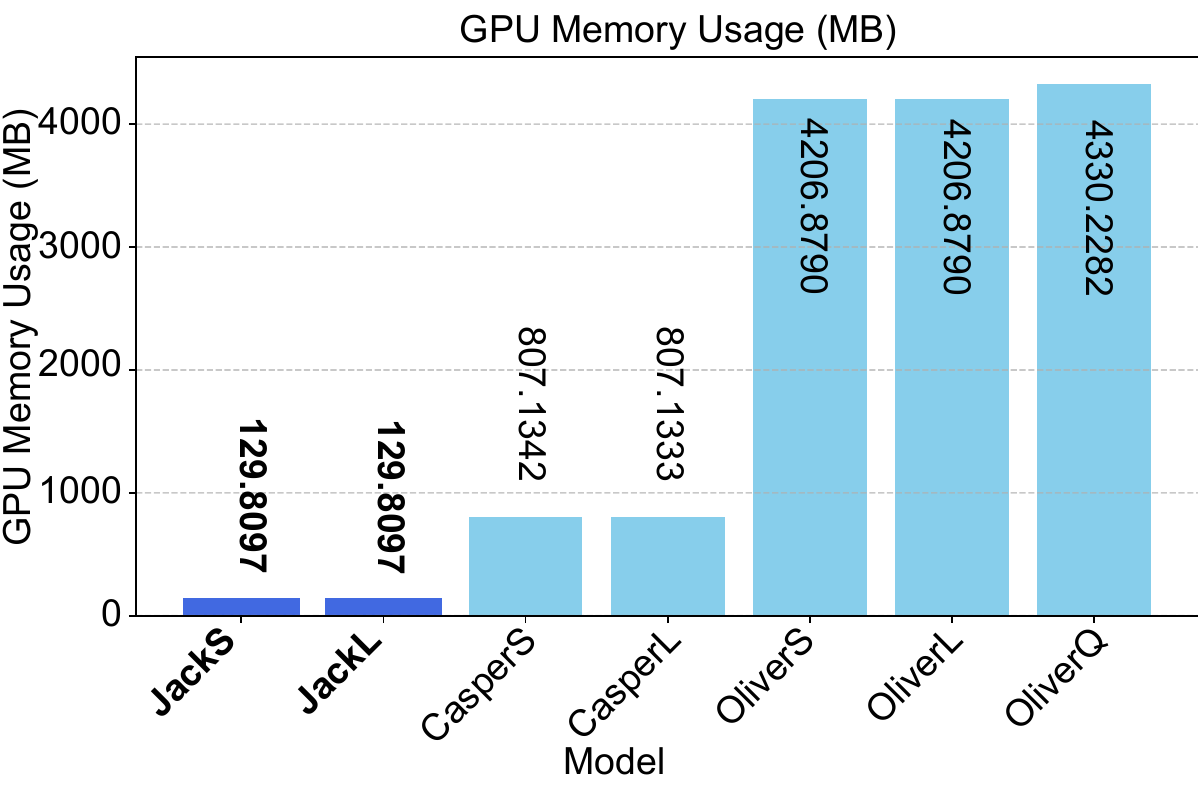} 
\caption{Average GPU memory consumption (VRAM in MB) during inference across different NPC model variants.}
\label{gpu-memory-usage-diagram}
\end{figure}
Figure~\ref{gpu-memory-usage-diagram} shows Jack variants had minimal GPU requirements ($\sim$130MB), ideal for low-resource setups. Casper ($\sim$807MB) balanced efficiency and performance, while Oliver variants ($\sim$4.2GB) remained feasible on higher-end consumer GPUs.

\subsubsection{Model Disk Footprint}
We measured disk storage requirements to evaluate deployment practicality.
\begin{figure}[h]
\centering
\includegraphics[width=.8\columnwidth]{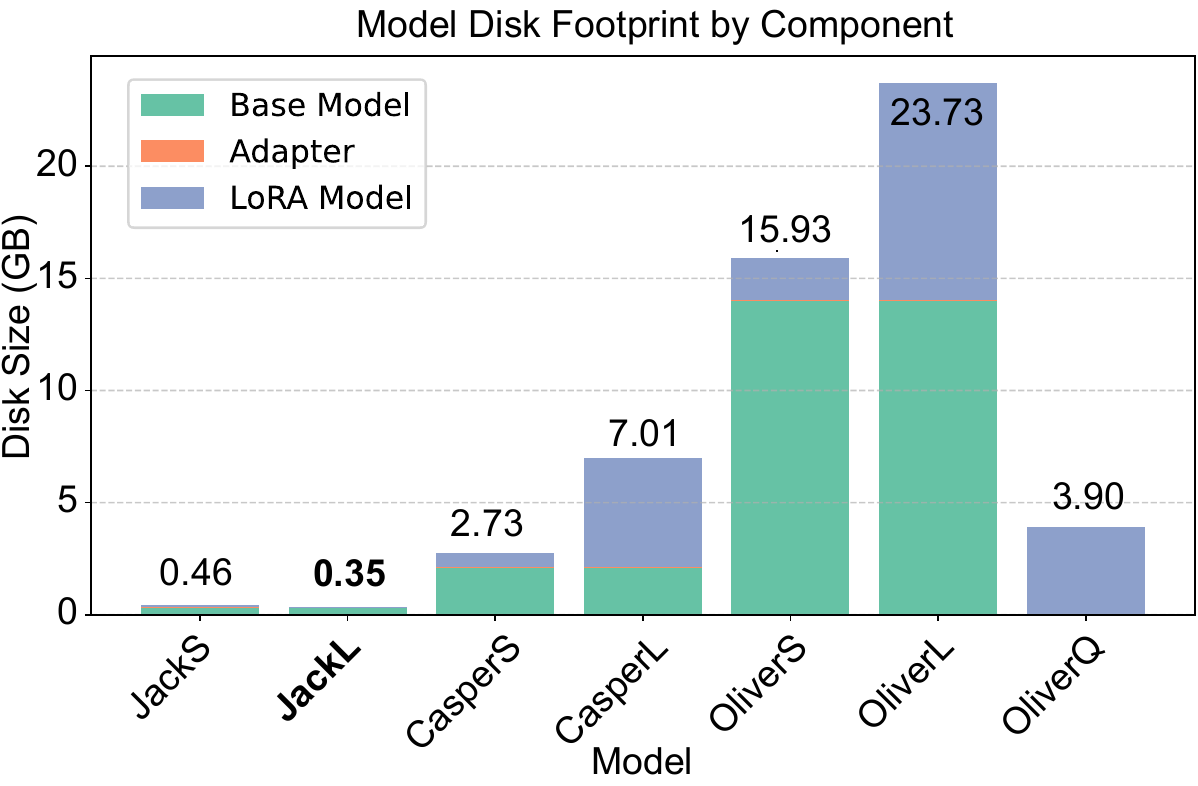} 
\caption{Disk storage footprint (GB) for each NPC model.}
\label{disk-footprint-diagram}
\end{figure}
Figure~\ref{disk-footprint-diagram} demonstrates that quantization significantly reduced OliverS footprint (OliverS 15.93GB, OliverQ 3,9GB). Jack ($\sim$0.4GB) and CasperS (2.73GB) variants also offered compact sizes, suitable for constrained environments.

\subsubsection{Latency (Total Response Generation Time)}
Latency was measured from input submission to full response delivery, crucial for real-time interactions.

\begin{figure}[h]
\centering
\includegraphics[width=.8\columnwidth, trim= 0 20 0 0,clip]{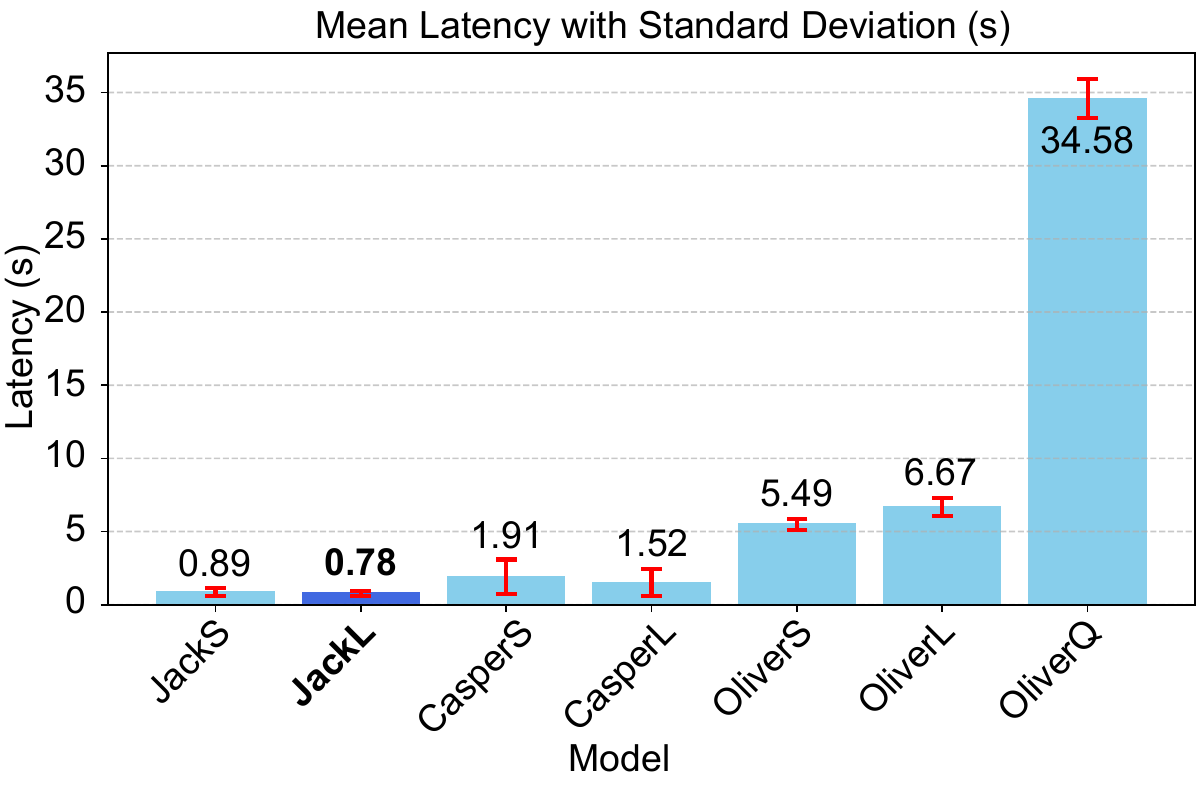} 
\caption{Mean latency (seconds) for response generation across models.}
\label{latency-mean-stddev-diagram}
\end{figure}

Figure~\ref{latency-mean-stddev-diagram} indicates Jack ($\sim$0.8s) and Casper ($\sim$1.7-1.9s) variants had low latency, suitable for interactive scenarios. OliverS's moderate latency (5.49s) remains acceptable with latency-masking methods (on-screen text rendering, Text-To-Speech), while OliverQ (34.58s) faced significant latency from quantization overhead.

\subsubsection{Time-To-First-Token (TTFT)}
Time-To-First-Token measures initial responsiveness from query to output generation start. An important metric to determine real-time suitability of the NPCs.

\begin{figure}[h]
\centering
\includegraphics[width=.8\columnwidth, trim= 0 35 0 0,clip]{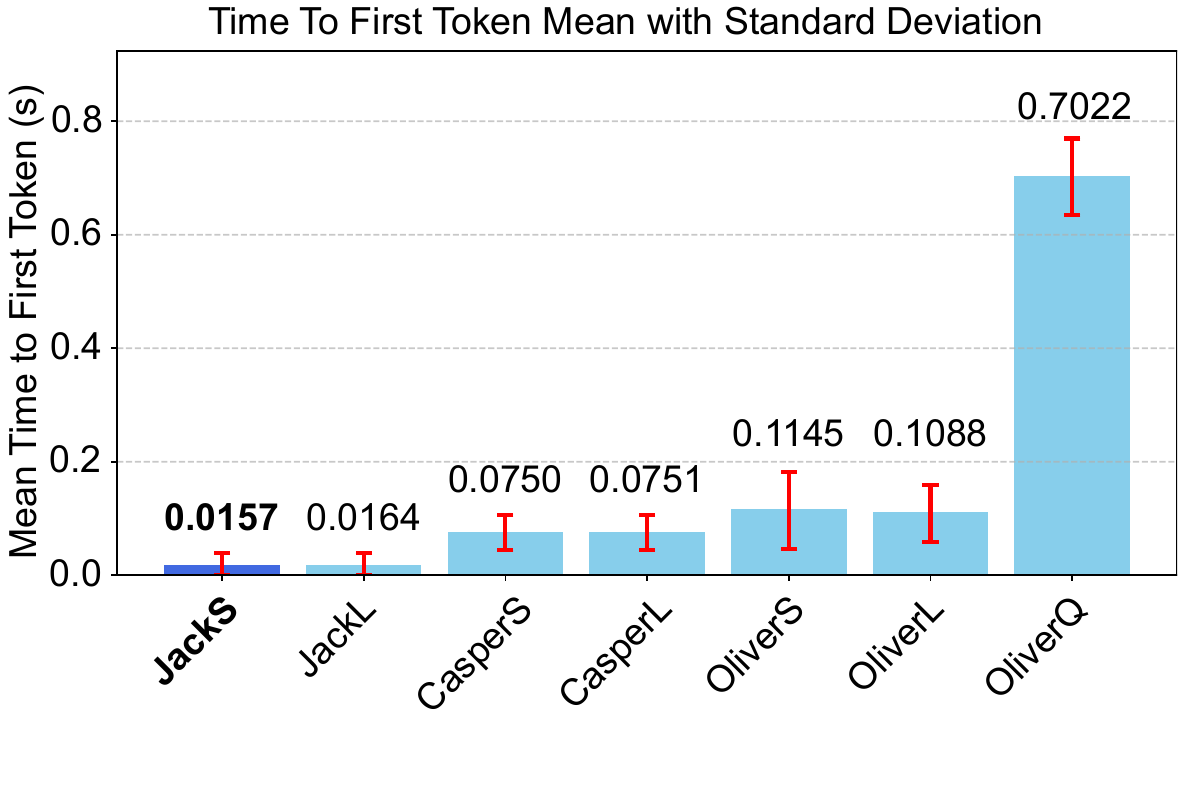} 
\caption{Mean time-to-first-token (TTFT in seconds) indicating initial model responsiveness after query submission.}
\label{ttft-mean-stddev-diagram}
\end{figure}

Figure~\ref{ttft-mean-stddev-diagram} shows all models except OliverQ had excellent initial responsiveness ($<$0.2s). OliverQ's notably higher TTFT (0.7022s) limits its immediate responsiveness in real-time settings.

\subsection{Runtime Modularity Metrics}

\subsubsection{Average Memory Swap Time}
We evaluated swap times between small (100 entries), medium (500), and large (1000) NPC memory databases, executing 50 swaps per combination. Test entries were standardized filler texts used exclusively for performance benchmarks.
\begin{figure}[h]
\centering
\includegraphics[width=.8\columnwidth, trim= 0 18 0 0,clip]{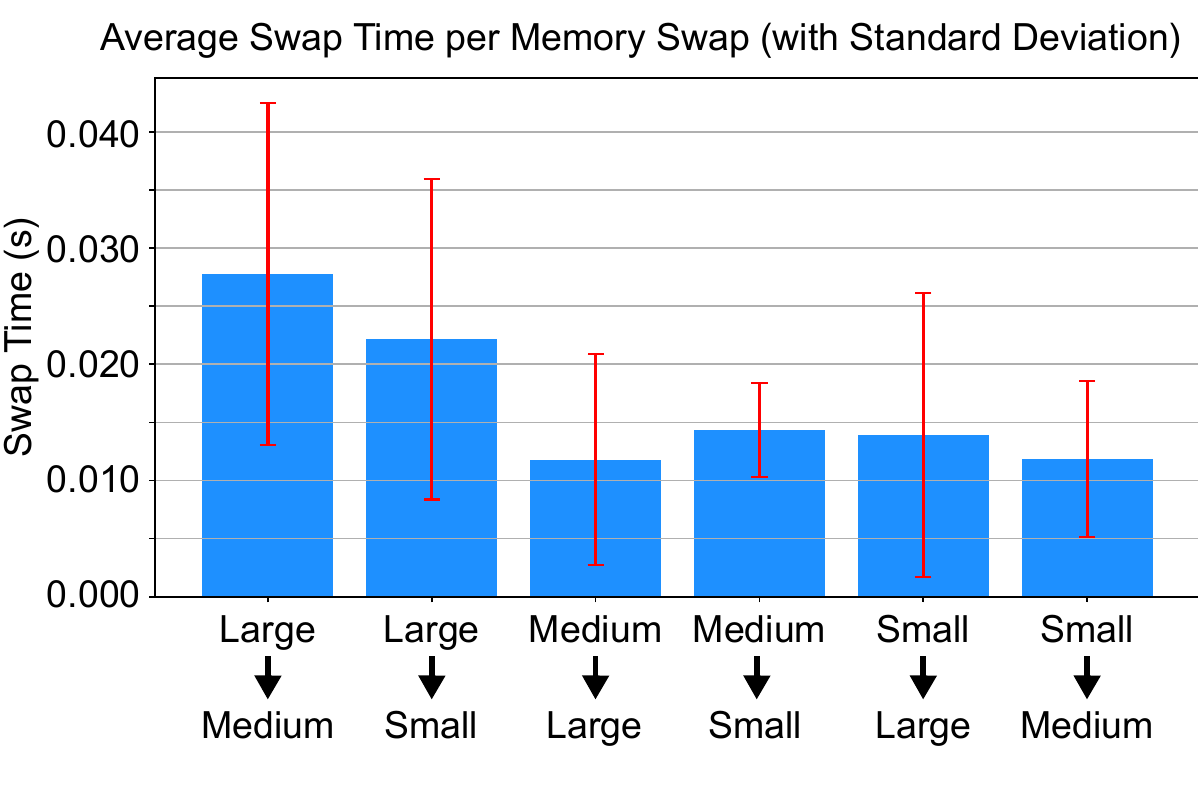} 
\caption{Average memory database swap time (seconds).}
\label{swap-time-mean-stddev-diagram}
\end{figure}
Figure~\ref{swap-time-mean-stddev-diagram} shows extremely fast memory swap times ($<$0.03s), enabling seamless and imperceptible NPC instance switching during gameplay. 

\subsubsection{Memory Retrieval Time}

Memory retrieval latency was evaluated by querying each database size (Small, Medium, Large) 50 times for both Conversational and World Knowledge entries.
\begin{figure}[h]
\centering
\includegraphics[width=.8\columnwidth, trim= 0 25 0 0,clip]{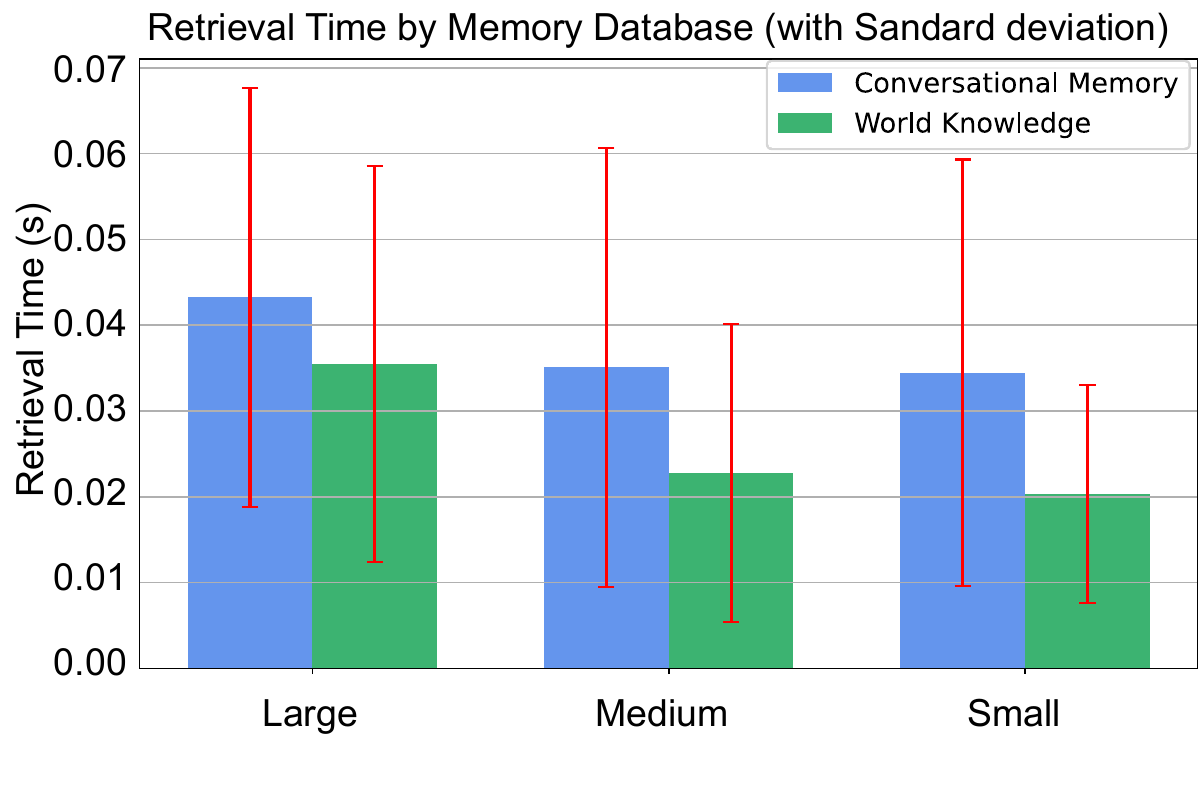} 
\caption{Average retrieval time (seconds) for entries from conversational history and world knowledge databases.}
\label{retrieval-time-mean-stddev-diagram}
\end{figure}
Figure~\ref{retrieval-time-mean-stddev-diagram} highlights very low retrieval latency ($<$0.042s), even for large databases, confirming system scalability and suitability for real-time player interactions.

\subsubsection{Memory Database Disk Footprint}
Disk footprints of memory databases were assessed across different entry counts to determine scalability.
\begin{figure}[h]
\centering
\includegraphics[width=.8\columnwidth]{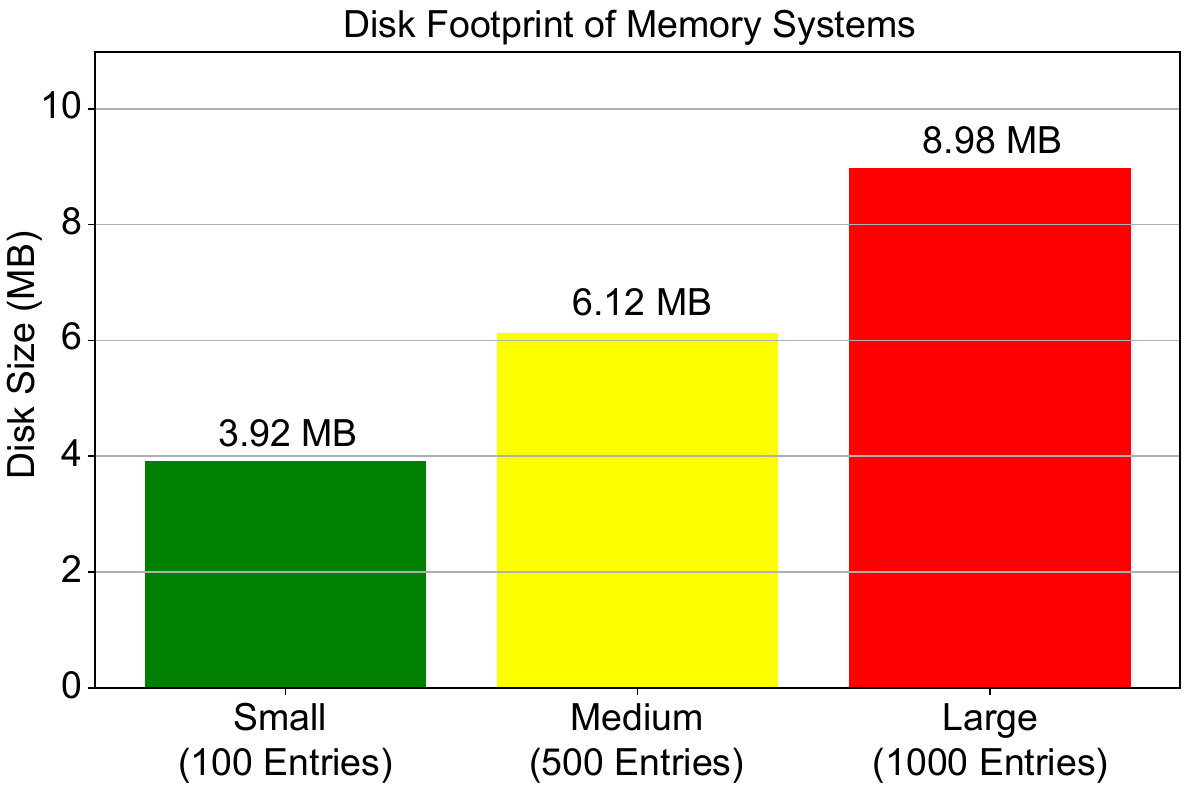} 
\caption{Disk footprint (MB) for small, medium, and large memory databases.}
\label{memory-disk-footprint-diagram}
\end{figure}
Figure~\ref{memory-disk-footprint-diagram} shows minimal footprint increases from small (3.92MB) to large (8.98MB), underscoring the modular memory system's practicality and scalability, enabling extensive and memory-rich NPC populations with minimal resource overhead.

Figure~\ref{memory-disk-footprint-diagram} illustrates minimal footprint growth (small 3.92MB, large 8.98MB), highlighting the practicality and efficiency of scalable modular memory storage.

\subsubsection{Note on RAM Usage during Swapping}
Due to ChromaDB's lazy loading and memory-mapped storage approach, explicit RAM usage measurements during memory swap were impractical. Data loading into RAM occurs only upon query access, maintaining memory efficiency. Consequently, this design minimizes RAM usage overhead during swaps, confirming the practicality and efficiency of our modular runtime memory architecture.
\section{Discussion}
\label{sec:discussion}

\subsection{Summary and Interpretation of Key Results}

Our evaluation demonstrates several clear insights regarding the effectiveness of small language models (SLMs) in powering modular NPC dialogue systems. Most notably, OliverS consistently achieved superior performance across all dialogue quality metrics, including factuality (93\%), context retention (100 \%), world knowledge retrieval (100 \%), and linguistic fluency (0 errors). This suggests that the Mistral-7B-Instruct model, when trained on a carefully curated smaller dataset, provides a robust balance of expressivity and accuracy, making it highly suitable for dialogue-rich, narrative-focused interactions.

The quantized variant, OliverQ, showed a substantially reduced disk footprint (from 15.93 GB to 3.9 GB), highlighting the effectiveness of quantization methods such as AutoGTPQ for storage efficiency. However, OliverQ exhibited significantly increased latency (34.58 seconds means latency compared to 5.49 seconds for OliverS), alongside moderate degradation in factual accuracy and context retention. This indicates that while quantization effectively reduces storage demands, it introduces non-trivial computational overhead, particularly noticeable on limited GPU hardware.

A consistent trend emerged regarding dataset size: smaller datasets (denoted by the suffix "S") generally outperformed their larger counterparts ("L") across factuality, context retention, world knowledge retrieval, and fluency. This counterintuitive result likely arises from decreased dataset quality or coherence in the larger synthetic dataset, introducing potential noise or conflicting examples. Additionally, larger datasets might lead to overfitting, negatively impacting generalization performance~\cite{brown2020languagemodelsfewshotlearners}. Consequently, our findings strongly suggest prioritizing smaller, high-quality datasets for NPC model fine-tuning, when using Low-Rank Adaptation (LoRA).

\subsection{Modularity and Runtime Memory Swapping}

A key contribution of this work is the demonstrated practical feasibility of a modular architecture employing runtime-swappable memory stores, allowing multiple distinct NPC interactions from a single fine-tuned base model. Our empirical validation of this approach provided compelling evidence for its practical viability:

Firstly, memory swap times were 
exceptionally low, ranging from 0.012 to 0.027 seconds across different scenarios (small, medium, and large databases). These negligible latencies confirm that memory swapping occurs practically seamlessly from the user's perspective, effectively supporting multiple NPC instances without perceptible gameplay disruptions.
Secondly, memory retrieval times remained consistently fast, even with the largest database size (1000 entries), yielding retrieval latencies below 0.042 seconds. This highlights the inherent scalability and responsiveness of the modular memory system, reinforcing its suitability for real-time gameplay contexts.
Importantly, memory databases showed minimal increases in disk footprint from small (3.92MB) to large (8.98MB), suggesting trivial storage overhead and strong scalability potential. This minimal increase makes our approach particularly well-suited for games requiring extensive NPC populations, each with their own distinct memory contexts.
Explicit RAM usage during memory swaps was not measurable due to ChromaDB's implementation of lazy loading and memory-mapped storage. This means data from a memory store is only loaded into RAM upon actual query access, 
providing significant practical advantages, maintaining RAM efficiency without unnecessary overhead.

Taken together, these results decisively confirm our hypothesis: runtime memory modularity is not only feasible but highly practical. A single fine-tuned NPC model can effectively power multiple unique NPC instances, each enriched by distinct, memory-driven interactions, enabling context-rich gameplay experiences.

\subsection{Trade-offs and Considerations for Model Selection}

Balancing dialogue quality and hardware performance is central to practical NPC deployment decisions. OliverS, based on the Mistral-7B-Instruct model fine-tuned with a smaller dataset, delivers exceptional dialogue quality suitable for high-importance NPCs demanding nuanced and contextually accurate interactions. However, its relatively higher latency (5.49 seconds) suggest practical deployment strategies involving latency masking through incremental text rendering or real-time text-to-speech (TTS) methods should be employed (as is the norm for LLMs)~\cite{ren2022fastspeech2fasthighquality}. The low Time-To-First-Token (0.1145 seconds) supports such approaches effectively, enhancing perceived responsiveness for real-time gameplay scenarios.

CasperS (TinyLlama) emerges as a balanced alternative combining good dialogue quality metrics (e.g., factuality 55\%, context retention 63.3\%, world knowledge retrieval 76.7\%) with reduced latency (1.91 seconds). Its moderate GPU memory usage ($\sim$807 MB) makes it highly suitable for broader NPC deployments with numerous responsive characters without sacrificing substantial dialogue quality.
Further experiments may substantiate CasperS's results.

The Jack models (DistilGPT-2), despite notably lower dialogue quality (factuality  $<$20\%, context retention $<$11\%), offer extremely low latency ($\sim$0.8 seconds) and minimal GPU usage ($\sim$130 MB). Such traits position Jack models as viable candidates for simple NPCs, but not for dialogue related scenarios.

The quantized OliverQ model presents complex trade-offs, significantly reducing disk footprint but suffering substantial latency penalties and moderate quality degradation. Its practical application thus hinges on scenarios prioritizing storage saving and deploying more capable hardware setups capable of managing increased latency. However, in any realistic scenario; A system capable of running a 4GB+ VRAM model will not have any issues with the $<$16GB disk space required to run OliverS, which outperforms OliverQ in every other metric except disk footprint.

\subsection{Practical Deployment Implications}

Our evaluation confirms the practicality of local deployment. GPU memory requirements ($\sim$4.2 GB for Oliver variants) are compatible with standard consumer-grade GPUs, making our NPC models accessible for most gaming setups.

Latency analysis highlights important usability considerations. OliverS's latency of 5.49 seconds is initially high but offset by an impressively low TTFT of 0.1145 seconds. By employing on-screen text rendering or TTS solutions, this latency can be effectively masked, significantly improving perceived responsiveness and ensuring real-time viability. CasperS provides an excellent compromise, with latency under 2 seconds suitable for wide-ranging NPC interactions without requiring additional masking strategies. For CasperS, further experimentation might improve dialogue related evaluation metrics. Jack models use less hardware resources, but this comes at a steep penalty in dialogue related metrics, and is therefore not recommended without careful consideration of its use-case.

The practicality of memory modularity further enhances our system's deployment appeal. Extremely low memory swap times (below 0.03 seconds) facilitate seamless NPC transitions, enabling game designers to efficiently populate extensive game worlds with diverse, memory-rich characters at minimal storage and computational cost.

Finally, while our approach is primarily motivated by applications within computer games, the modular, persona-driven memory architecture has significant potential for broader deployment in other domains. Applications such as virtual assistants, customer support bots, interactive educational systems, or any scenario requiring multiple expressive conversational agents with distinct long-term memory stores could readily benefit from our proposed approach.

\section{Conclusion}
\label{sec:conclusion}

This paper introduced a novel modular dialogue architecture for non-playable characters (NPCs) using LoRA fine-tuned Small Language Models (SLMs) paired with runtime-swappable memory modules. Our primary contribution lies in demonstrating that a single fine-tuned base model can efficiently power multiple distinct NPC instances, each enriched by unique conversational and world-knowledge memories. This approach significantly reduces computational and storage overhead, enabling expressive and scalable systems suitable for real-time deployment in modern games. While primarily designed for NPC dialogue in games, the architecture's modularity and efficiency also suggest promising applicability to other domains requiring memory-rich conversational agents.

Through comprehensive evaluations, we showed that the Mistral-7B-Instruct model fine-tuned with a carefully curated smaller synthetic dataset (OliverS) consistently provided superior dialogue quality, including near-perfect factual accuracy, context retention, and linguistic fluency. Although quantization dramatically reduced the disk footprint, it introduced substantial latency penalties, highlighting important trade-offs. Additionally, our results highlighted the 
viability of memory modularity, demonstrating negligible memory swap times and excellent retrieval scalability.

In practical terms, our architecture supports seamless real-time interactions through effective latency-masking techniques such as incremental text rendering or real-time text-to-speech. The modular memory approach ensures minimal overhead, making it highly suitable for games featuring extensive and diverse NPC populations, all powered by a single fine-tuned model.

Our work also identifies several important limitations that suggest clear directions for future research: 

\begin{compactitem}
    \item Since dataset quality emerged as a critical factor for NPC performance, future work should explore improving dataset quality/refinement methods. Perhaps through iterative feedback loops~\cite{ouyang2022traininglanguagemodelsfollow}.
    \item Latency penalties introduced by quantization require more exploration of alternative methods or optimized quantization techniques~\cite{dettmers2022llmint8, stock2020bitgoesdownrevisiting}.
    \item Current system uses static personas, future work could explore runtime persona adjustments~\cite{see2019makesgoodconversationcontrollable, dinan2019wizardwikipediaknowledgepoweredconversational}.
    \item Future work should explore real-world player feedback studies, to see if SLM powered NPCs could enhance player experience~\cite{roller2020recipesbuildingopendomainchatbot, Ji_2023}.
\end{compactitem}

\appendix

\bibliography{aaai2026}

@inproceedings{gao-emami-2023-turing,
    title = "The {T}uring Quest: Can Transformers Make Good {NPC}s?",
    author = "Gao, Qi Chen  and
      Emami, Ali",
    editor = "Padmakumar, Vishakh  and
      Vallejo, Gisela  and
      Fu, Yao",
    booktitle = "Proceedings of the 61st Annual Meeting of the Association for Computational Linguistics (Volume 4: Student Research Workshop)",
    month = jul,
    year = "2023",
    address = "Toronto, Canada",
    publisher = "Association for Computational Linguistics",
    url = "https://aclanthology.org/2023.acl-srw.17/",
    doi = "10.18653/v1/2023.acl-srw.17",
    pages = "93--103"
}

@misc{sarahparvinicanaimakegamesmoreimmersive,
      title={Can AI make video games more immersive? Some studios turn to AI-fueled NPCs for more interaction}, 
      author={Sarah Parvini},
      year={2024},
      url={https://www.ap.org/news-highlights/spotlights/2024/can-ai-make-video-games-more-immersive-some-studios-turn-to-ai-fueled-npcs-for-more-interaction/}, 
    note = {Accessed: 19-03-2025}
}

@inproceedings{schlunder2013greetings,
  title={Greetings Generation in Video Role Playing Games},
  author={Schl{\"u}nder, Bj{\"o}rn and Klabunde, Rainer},
  booktitle={Proceedings of the 14th European Workshop on Natural Language Generation},
  pages={167--171},
  year={2013},
  organization={Association for Computational Linguistics},
  url={https://aclanthology.org/W13-2122.pdf}
}

@misc{languagetool,
  author = {{LanguageTool}},
  year = {2025},
  title = {LanguageTool: Open-source grammar, style, and spell checker},
  howpublished = {\url{https://languagetool.org/}},
  note = {Accessed: 2025-05-24}
}

@misc{nvidia2023ace,
  author = {{NVIDIA Corporation}},
  title = {NVIDIA ACE for Games: Generative {AI} NPCs},
  howpublished = {\url{https://www.nvidia.com/en-us/geforce/news/nvidia-ace-for-games-generative-ai-npcs/}},
  year = {2023},
  note = {Accessed: 2025-03-18}
}

@misc{sanh2020distilbertdistilledversionbert,
      title={DistilBERT, a distilled version of BERT: smaller, faster, cheaper and lighter}, 
      author={Victor Sanh and Lysandre Debut and Julien Chaumond and Thomas Wolf},
      year={2020},
      eprint={1910.01108},
      archivePrefix={arXiv},
      primaryClass={cs.CL},
      url={https://arxiv.org/abs/1910.01108}, 
}

@misc{distilgpt2,
  author = {{DistilBERT}},
  title = {HuggingFace Page for DistilGPT2},
  howpublished = {\url{https://huggingface.co/distilbert/distilgpt2}},
  year = {2023},
  note = {Accessed: 19-03-2025}
}

@misc{zhang2024tinyllamaopensourcesmalllanguage,
      title={TinyLlama: An Open-Source Small Language Model}, 
      author={Peiyuan Zhang and Guangtao Zeng and Tianduo Wang and Wei Lu},
      year={2024},
      eprint={2401.02385},
      archivePrefix={arXiv},
      primaryClass={cs.CL},
      url={https://arxiv.org/abs/2401.02385}, 
}

@misc{tinyllama2023,
  author = {{TinyLlama Team}},
  title = {TinyLlama-1.1B-Chat-v1.0},
  howpublished = {\url{https://huggingface.co/TinyLlama/TinyLlama-1.1B-Chat-v1.0}},
  year = {2023},
  note = {Accessed: 19-03-2025}
}

@misc{jiang2023mistral7b,
      title={Mistral 7B}, 
      author={Albert Q. Jiang and Alexandre Sablayrolles and Arthur Mensch and Chris Bamford and Devendra Singh Chaplot and Diego de las Casas and Florian Bressand and Gianna Lengyel and Guillaume Lample and Lucile Saulnier and Lélio Renard Lavaud and Marie-Anne Lachaux and Pierre Stock and Teven Le Scao and Thibaut Lavril and Thomas Wang and Timothée Lacroix and William El Sayed},
      year={2023},
      eprint={2310.06825},
      archivePrefix={arXiv},
      primaryClass={cs.CL},
      url={https://arxiv.org/abs/2310.06825}, 
}

@misc{ikennoli2024deploythefirstondeviceslmforimprovedgamecharacterroleplay,
    title={Deploy the First On-Device Small Language Model for Improved Game Character Roleplay}, 
    author={Ike Nnoli},
    year={2024},
    url={https://developer.nvidia.com/blog/deploy-the-first-on-device-small-language-model-for-improved-game-character-roleplay/#:~:text=Nemotron,LLMs},
    note = {Accessed: 19-03-2025}
}

@misc{hu2021lora,
  title={LoRA: Low-Rank Adaptation of Large Language Models},
  author={Edward J. Hu and Yelong Shen and Phillip Wallis and Zeyuan Allen-Zhu and Yuanzhi Li and Shean Wang and Weizhu Chen},
  year={2021},
  eprint={2106.09685},
  archivePrefix={arXiv},
  primaryClass={cs.LG}
}

@misc{houlsby2019parameterefficienttransferlearningnlp,
      title={Parameter-Efficient Transfer Learning for NLP}, 
      author={Neil Houlsby and Andrei Giurgiu and Stanislaw Jastrzebski and Bruna Morrone and Quentin de Laroussilhe and Andrea Gesmundo and Mona Attariyan and Sylvain Gelly},
      year={2019},
      eprint={1902.00751},
      archivePrefix={arXiv},
      primaryClass={cs.LG},
      url={https://arxiv.org/abs/1902.00751}, 
}

@misc{mistral2023,
  author = {{Mistral AI}},
  title = {Mistral-7B-v0.1},
  howpublished = {\url{https://huggingface.co/mistralai/Mistral-7B-v0.1}},
  year = {2023},
  note = {Accessed: 19-03-2025}
}

@misc{reimers2019sentencebertsentenceembeddingsusing,
      title={Sentence-BERT: Sentence Embeddings using Siamese BERT-Networks}, 
      author={Nils Reimers and Iryna Gurevych},
      year={2019},
      eprint={1908.10084},
      archivePrefix={arXiv},
      primaryClass={cs.CL},
      url={https://arxiv.org/abs/1908.10084}, 
}

@misc{chromadb2023,
  author       = {{Chroma Team}},
  title        = {Chroma: The Open-Source Embedding Database},
  year         = {2023},
  howpublished = {\url{https://www.trychroma.com}},
  note         = {Accessed: 2024-03-25}
}

@misc{wang2024openchatadvancingopensourcelanguage,
      title={OpenChat: Advancing Open-source Language Models with Mixed-Quality Data}, 
      author={Guan Wang and Sijie Cheng and Xianyuan Zhan and Xiangang Li and Sen Song and Yang Liu},
      year={2024},
      eprint={2309.11235},
      archivePrefix={arXiv},
      primaryClass={cs.CL},
      url={https://arxiv.org/abs/2309.11235}, 
}

@misc{frantar2023gptqaccurateposttrainingquantization,
      title={GPTQ: Accurate Post-Training Quantization for Generative Pre-trained Transformers}, 
      author={Elias Frantar and Saleh Ashkboos and Torsten Hoefler and Dan Alistarh},
      year={2023},
      eprint={2210.17323},
      archivePrefix={arXiv},
      primaryClass={cs.LG},
      url={https://arxiv.org/abs/2210.17323}, 
}

@misc{brown2020languagemodelsfewshotlearners,
      title={Language Models are Few-Shot Learners}, 
      author={Tom B. Brown and Benjamin Mann and Nick Ryder and Melanie Subbiah and Jared Kaplan and Prafulla Dhariwal and Arvind Neelakantan and Pranav Shyam and Girish Sastry and Amanda Askell and Sandhini Agarwal and Ariel Herbert-Voss and Gretchen Krueger and Tom Henighan and Rewon Child and Aditya Ramesh and Daniel M. Ziegler and Jeffrey Wu and Clemens Winter and Christopher Hesse and Mark Chen and Eric Sigler and Mateusz Litwin and Scott Gray and Benjamin Chess and Jack Clark and Christopher Berner and Sam McCandlish and Alec Radford and Ilya Sutskever and Dario Amodei},
      year={2020},
      eprint={2005.14165},
      archivePrefix={arXiv},
      primaryClass={cs.CL},
      url={https://arxiv.org/abs/2005.14165}, 
}

@misc{ren2022fastspeech2fasthighquality,
      title={FastSpeech 2: Fast and High-Quality End-to-End Text to Speech}, 
      author={Yi Ren and Chenxu Hu and Xu Tan and Tao Qin and Sheng Zhao and Zhou Zhao and Tie-Yan Liu},
      year={2022},
      eprint={2006.04558},
      archivePrefix={arXiv},
      primaryClass={eess.AS},
      url={https://arxiv.org/abs/2006.04558}, 
}

@misc{ouyang2022traininglanguagemodelsfollow,
      title={Training language models to follow instructions with human feedback}, 
      author={Long Ouyang and Jeff Wu and Xu Jiang and Diogo Almeida and Carroll L. Wainwright and Pamela Mishkin and Chong Zhang and Sandhini Agarwal and Katarina Slama and Alex Ray and John Schulman and Jacob Hilton and Fraser Kelton and Luke Miller and Maddie Simens and Amanda Askell and Peter Welinder and Paul Christiano and Jan Leike and Ryan Lowe},
      year={2022},
      eprint={2203.02155},
      archivePrefix={arXiv},
      primaryClass={cs.CL},
      url={https://arxiv.org/abs/2203.02155}, 
}

@article{dettmers2022llmint8,
  title={LLM.int8(): 8-bit Matrix Multiplication for Transformers at Scale},
  author={Tim Dettmers and Mike Lewis and Younes Belkada and Luke Zettlemoyer},
  journal={arXiv preprint arXiv:2208.07339},
  year={2022}
}

@misc{stock2020bitgoesdownrevisiting,
      title={And the Bit Goes Down: Revisiting the Quantization of Neural Networks}, 
      author={Pierre Stock and Armand Joulin and Rémi Gribonval and Benjamin Graham and Hervé Jégou},
      year={2020},
      eprint={1907.05686},
      archivePrefix={arXiv},
      primaryClass={cs.CV},
      url={https://arxiv.org/abs/1907.05686}, 
}

@misc{see2019makesgoodconversationcontrollable,
      title={What makes a good conversation? How controllable attributes affect human judgments}, 
      author={Abigail See and Stephen Roller and Douwe Kiela and Jason Weston},
      year={2019},
      eprint={1902.08654},
      archivePrefix={arXiv},
      primaryClass={cs.CL},
      url={https://arxiv.org/abs/1902.08654}, 
}

@misc{dinan2019wizardwikipediaknowledgepoweredconversational,
      title={Wizard of Wikipedia: Knowledge-Powered Conversational agents}, 
      author={Emily Dinan and Stephen Roller and Kurt Shuster and Angela Fan and Michael Auli and Jason Weston},
      year={2019},
      eprint={1811.01241},
      archivePrefix={arXiv},
      primaryClass={cs.CL},
      url={https://arxiv.org/abs/1811.01241}, 
}

@misc{roller2020recipesbuildingopendomainchatbot,
      title={Recipes for building an open-domain chatbot}, 
      author={Stephen Roller and Emily Dinan and Naman Goyal and Da Ju and Mary Williamson and Yinhan Liu and Jing Xu and Myle Ott and Kurt Shuster and Eric M. Smith and Y-Lan Boureau and Jason Weston},
      year={2020},
      eprint={2004.13637},
      archivePrefix={arXiv},
      primaryClass={cs.CL},
      url={https://arxiv.org/abs/2004.13637}, 
}

@article{Ji_2023,
   title={Survey of Hallucination in Natural Language Generation},
   volume={55},
   ISSN={1557-7341},
   url={http://dx.doi.org/10.1145/3571730},
   DOI={10.1145/3571730},
   number={12},
   journal={ACM Computing Surveys},
   publisher={Association for Computing Machinery (ACM)},
   author={Ji, Ziwei and Lee, Nayeon and Frieske, Rita and Yu, Tiezheng and Su, Dan and Xu, Yan and Ishii, Etsuko and Bang, Ye Jin and Madotto, Andrea and Fung, Pascale},
   year={2023},
   month=mar, pages={1–38} 
}

\newpage

\end{document}